# GLOCON Database: Design Decisions and User Manual (v1.0)

Ali Hürriyetoğlu (ahurriyetoglu@ku.edu.tr), Osman Mutlu (omutlu@ku.edu.tr), Fırat Duruşan (fdurusan@ku.edu.tr), Erdem Yörük (eryoruk@ku.edu.tr)

*Date*: May 28, 2024

## Table of Contents

## 1. Introduction

GLOCON is a database of contentious events automatically extracted from national news sources from various countries in multiple languages. National news sources are utilized, and complete news archives are processed to create an event list for each source. Automation is achieved using a gold standard corpus sampled randomly from complete news archives (Yörük et al. 2022) and all annotated by at least two domain experts based on the event definition provided in Duruşan et al. (2022).

The database consists of the following countries and sources provided in Table 1 as of May 2024.

*Table 1 Summary table of GLOCON as of May, 2024.*

|  | News Source | Period | Processed Language | Event Count |
|---|---|---|---|---|
| Argentina | La Nacion | 1995-12-19 : 2021-12-31 | Spanish | 68,281 |
| Brasil | Estadao | 2000-05-16 : 2022-12-01 | Portuguese | 28,974 |
| Brasil | Folha | 1989-11-19 : 2022-10-16 | Portuguese | 78,178 |
| India | The Hindu | 2006-01-01 : 2021-09-14 | English | 459,539 |
| South Africa | Sabinet | 2002-11-04 : 2019-09-22 | English | 45,951 |

GLOCON is created using the code repository on https://github.com/emerging-welfare/emw_pipeline_nf. Please see https://glocon.ku.edu.tr/ for additional information and an online database demo.

The last version of this document and any updates will be communicated using the repository https://github.com/emerging-welfare/glocon-release. You should follow the procedure specified on https://glocon.ku.edu.tr to obtain the data.

This release contains a manually created events list for Turkey as well. The definition of an event is the same as the rest of the database. Any questions about this part should be specifically communicated to Prof. Erdem Yörük.

This document provides all relevant information to ensure the best use of the GLOCON database.

## 2. Contents of GLOCON Database

The data is released as JSON files organized as one separate file for each news source. Each JSON file consists of JSON objects, one per each line, considered as .jsonl. Each line, which is an event record, contains various information types about an event. An annotated sample of an event record is provided below.

```
{'coordinate_dates': '2021',   → The geocoding reference, which may differ in relation to official place names in a new country.
 'day': 21,   → Publication date of the news article.
 'district_name': 'Berazategui', ->  District name as a result of the geocoding step.
 'event_id': 'lanacion_2022-12-08_00000001',   → A unique ID we assigned by the Project team.
 'event_sentence_numbers': [16],   → Event sentence number as it occur in the original news text. This one was the 16th sentence extracted from the respective article.
```

```
 'event_sentences': ['Berazategui: un importante incendio tomó gran parte
del Parque Pereyra Temas Hoy Santa Fe Salud Conforme a os criterios de Con
océ The Trust Project'],  → *Event sentence as it occurs in the news article.*
 'eventcategory': 'Demonstration',  → *Automatically predicted event category.*
 'latitude': -34.7632391323248,  → *Latitude of the event place as identified in the geocoding step.*
 'longitude': -58.1150116488357,  → *Longitude of the event place as ideintified in the geocoding step.*
 'month': 12,  → *Month of the news article publication date.*
 'organizer0category': 'Unknown',  → *Automatically predicted organizer category using text classification.*
 'organizers': [],  → *Automatically extracted organizers in the event sentence(s).*
 'participant0category': 'Unknown',  → *Automatically predicted participant category using text classification.*
 'participants': [],  → *Automatically extracted participants in the event sentence(s).*
 'state_name': 'Buenos Aires',  → *State name of the event place.*
 'targets': [],  → *Automatically extracted targets in the event sentence(s).*
 'text_snippet': 'Una madre de 50 años fue autorizada por la Justicia de
Familia de Rosario, provincia de Santa Fe, a someterse al procedimiento de
fecundación asistida para gestar en su vientre a su nieto o nieta . Su hij
a, de 25 años, fue diagnosticada con un síndrome que no le permite embaraz
arse por carecer de útero y ',  → *First part of a news article.*
 'title': 'Una mujer de 50 años gestará en su vientre a su nieto porque su
hija no tiene útero. El procedimiento se hará por fecundación asistida; el
embrión corresponde a la joven y su pareja',  → *The title of the news article.*
 'triggers': ['incendio'],  → *Automatically detected event trigger.*
 'url': 'https://www.lanacion.com.ar-d3f4g5-sociedad-d3f4g5-una-mujer-de-5
0-anos-gestara-en-su-vientre-a-su-nieto-porque-su-hija-no-tiene-utero-nid2
1122021-d3f4g5-',  → URL of the news article
 'violent': 'non-violent',  → *Automatically predicted label for an event.*
 'year': 2021  → *Year of the news article publication date.*
}
```

Detailed information about event records and how they are created are provided in the following subsections.

### 2.1 Events

The event definition and the annotation manual are described in Duruşan et al. (2022).

We developed a multitask multilingual model that generates predictions for four tasks described below (Mutlu and Hürriyetoğlu, 2023; Hürriyetoğlu 2021a).

1- *Document classification*: It classifies a news article as relevant and irrelevant (containing event or not).
2- *Sentence classification*: It classifies a sentence in a news article as containing event or not.
3- *Event sentence coreference identification (ESCI)*: Sentences that are about the same event are grouped together (More information can be found as ESCI task in Hürriyetoğlu et al. (2020), Subtask 3 in Hürriyetoğlu et al. (2021), and Subtask 3 in Hürriyetoğlu et al. (2022b)). This task is handled as a traditional event coreference task in (Hürriyetoğlu et al. 2022a).
4- *Event extraction*: Event trigger, time, place, etc. are identified.

An event record is constructed from event information extracted from a group of event sentences (a group consists of one or more event sentences) referring to same event. The grouping is achieved using ESCI techniques specified in step 3 above. There may be multiple event records from a relevant news article.

Place name is extracted both from event sentences and from an HTML document of the news article. If a place name extracted in at least one of the sentences in an event sentence group this is used as the event place. Otherwise, the place name extracted from the HTML document is used.

Event time is utilized as the publication date of a news article mentioning an event.

### 2.1.1 Semantic category

An event can be one of the following types: "Demonstration", "Armed Militancy", "Group Clash", "Industrial Action", "Other". Some indicative performance indicators for this classification can be found in the Section 6.1 of Hürriyetoğlu et al. (2021b).

### 2.1.2 Violent events

The field 'violent' provides the result of a binary text classifier that processes an entire news article classified as containing a relevant event and predicts whether it contains a violent event. If the event is violent, the value of this field is 'violent' otherwise 'non-violent'. Document level.

## 2.2 Geocoding

The geocoding is performed using the code (https://github.com/emerging-welfare/emw_pipeline_nf/tree/multi_task/bin/scripts) and resources https://github.com/emerging-welfare/geocoding_dictionaries (resources) created specifically for this project. This code considers place name changes into account. A place name is interpreted according to the publication date of a news article. The field 'coordinate_dates' in an event record indicates the respective file for the coordinates utilized for assigning latitude and longitude for an event.

All this information is merged using the code in https://github.com/emerging-welfare/emw_pipeline_nf/blob/multi_task/bin/scripts/construct_event_database.py.

If both the latitude and longitude are '0' in an event record, this means the geocoding was not successful in determining these values. These cases contain district and state names most of the time.

An event is included in an event list if the place of the event can be determined, and the place name can be geocoded. A place name can be either publication place or the place name that occur in a event sentence group. In case i) the event extraction model does not detect a place name in any of the event sentences in a group; ii) none of the place names detected geocoded, this sentence group is excluded from the event list; or iii) If a place name is not in the country where the news article is published, this event record is excluded from an event list extracted for a country. Therefore,

## 2.3 Organizer, Participant, Target detection

We create two different information for organizer, participant, and target information. The first one is a machine learning model based on text classification. These classifiers (two, a separate one for organizer and one for participant types) process event sentence groups and predict a label. The labels are:

Organizer: "Political Party", "Grassroots Organization", "Labor Union", "Militant Organization", "Chambers of Professionals", "No"

Participant: "Peasant", "Worker", "Professional", "Student", "People", "Politician", "Activist", "Militant", "Other", "No"

The event records encode the prediction 'no' in the Organizer and Participant types as 'Unknown'. This is an explicitly encoded label in these classifiers.

The organizer(s) and participant(s) categories are provided in the 'organizer0category' and 'participant0category'. These categories may be more than one and encoded as 'organizer1category', 'participant1category', etc., in case event sentences yield different predictions.

The second information type is the output of the event extraction model. Organizer, participant, and target information is extracted as it occurs in the event sentences. This information is provided by the field *organizers*, *participants*, and *targets*. A sample is provided below.
    'participants': ['agro', 'vándalos', 'delincuentes', 'autores']

If our tools do not detect relevant information, the fields related to organizer, participant, and target may be missing.

## 2.4 Castes in India

We developed a rule-based classifier that identifies caste group of an event in India. The caste groups are … This information is specific to India and encoded in a field … .

## 2.5 Women related events in India

We developed a rule-based classifier that identifies events related to Women in India. This is a binary classification and based on keywords related to women and associations related to women.

# 3  Files in a release

The project team shares a folder with subfolders for each country and each source utilized for creating an event list for that country.

Offline copies of the source code and resources are provided in the online folder shared as well.

## 4 Additional Information

The standard files in the release contain summary or aggregated information. If you need raw information such as raw text, sentences extracted from text, event sentence groups, results of the geocoding steps, URL of the news article, or any debugging information, please contact us.

A gold standard corpus, which we name as GLOCON Gold (https://github.com/emerging-welfare/glocongold), is created for developing and validating the automated tools. This corpus is available per request as well.

The data is provided to you as a result of your application and only for research purposes. You should not share this data under any condition.

As most of the steps are automated, errors may be encountered in the database. You accept these inaccuracies.

### 4.1 Contacting us

Use the contact information on the <release repo> or respective repo if you have a specific question. We would appreciate any insight or feedback about the database or the documentation.

There is an email list automated-political-event-collection@googlegroups.com.

Twitter account: @EmergingWelfare

Youtube account: https://www.youtube.com/@emergingwelfareproject8979

You can request the data by reaching the authors or, preferably, first complete the form on https://glocon.ku.edu.tr/dataset/ and next notify the authors.